\documentclass{ifacconf}

\usepackage{graphicx}
\usepackage[numbers,sort&compress]{natbib}
\usepackage{threeparttable}

\usepackage{colortbl}
\usepackage{listings}
\usepackage{amsmath}
\usepackage{multirow}
\usepackage{xurl}
\begin{document}
\begin{frontmatter}

\title{LLM-Assisted Planning with Distributed Onboard Behaviour Tree Execution for Multi-AUV Missions}
\thanks[copyrightnotice]{\copyright\ 2026 the authors. This work has been accepted to IFAC for publication under a Creative Commons Licence CC-BY-NC-ND.}

\author[First]{Ehsan Khorrambakht}
\author[First]{Cristiano Caissutti}
\author[First]{Andrea Munafo}
\author[First]{Andrea Caiti}

\address[First]{Department of Information Engineering, University of Pisa \\ Pisa, Italy \\ (corresponding authors: ehsan.khorrambakht@ing.unipi.it, cristiano.caissutti@phd.unipi.it)}

\begin{abstract}
Coordinating autonomous underwater vehicles requires translating operator intent into feasible assignments while preserving autonomous execution when communication is limited. We present a hierarchical framework that couples Large Language Model (LLM) interpretation of natural-language mission instructions with Mixed-Integer Linear Programming (MILP) and distributed Behaviour Tree (BT) execution. The LLM extracts assignment and ordering constraints, which the MILP optimizer combines with vehicle capabilities and temporal requirements. A BT--LLM Bridge distributes the resulting allocations to onboard BT Mission Managers. Per-vehicle Robotic Agents assess local faults and request selective replanning, allowing unaffected vehicles to continue their assigned tasks. Evaluation in a ROS-based AUV simulator compares Qwen-3-8B and Llama-3.1-8B on constraint extraction and tests recovery from sensor degradation and critical vehicle faults. Qwen-3-8B provides more accurate constraint extraction, and all tasks are completed in the tested failure scenarios. These simulation results support the feasibility of combining language-based mission specification with structured allocation and distributed execution.
\end{abstract}


\end{frontmatter}

\section{Introduction}
In recent years, the growing deployment of autonomous vehicles in maritime environments has emphasized the need for trustworthy autonomy frameworks that enhance human supervision and ensure reliable mission execution. These environments are dynamic, uncertain, and characterized by bandwidth-limited communication, causing single vehicles to struggle due to limited endurance, sensing range, and communication capabilities. Cooperative fleets of Autonomous Underwater Vehicles (AUVs) can mitigate these limitations, yet coordinating multiple vehicles remains challenging under intermittent acoustic connectivity \citep{stojanovic2009underwater}.

Large Language Models (LLMs) offer new capabilities for interpreting natural-language (NL) mission specifications and generating structured plans. However, they introduce risks for safety-critical underwater operations, including infeasible outputs, limited grounding in robot capabilities, and inference latency incompatible with real-time control \citep{liang2022code, ahn2022saycan}. Behaviour Trees (BTs), by contrast, provide deterministic, modular, and interpretable mission execution \citep{sprague2018petter}. Recent efforts have explored LLM-assisted BT generation \citep{izzo2024btgenbot}, but these approaches focus on single robots and assume reliable communication.

Despite progress in LLM-based planning, BT execution, and multi-AUV coordination, three challenges remain. First, existing systems do not support NL interfaces for multi-robot task allocation involving heterogeneous capabilities and task dependencies, still relying on structured mission files \citep{wang2023auv}. Second, distributed failure handling under intermittent acoustic communication is limited. Third, current LLM--BT integrations lack mechanisms to align high-level LLM reasoning with the deterministic execution required across multiple vehicles.

This work addresses these gaps through a hierarchical architecture that integrates LLM-assisted planning with distributed BT execution for robust multi-AUV missions. The main contributions are as follows:
\begin{enumerate}
    \item An LLM-assisted Planner which extracts operator constraints and embeds them in a Mixed-Integer Linear Program (MILP) for capability-aware task allocation.
    \item Distributed Robotic Agents (\emph{RAs}) enabling autonomous failure assessment and selective replanning, reducing communication demands.

    \item A BT Mission Manager on each vehicle enabling deterministic, modular, and reactive onboard execution of navigation and survey tasks using pre-defined BT subtrees.
    \item A topside--seaside separation in which all semantic interpretation and multi-robot reasoning occur topside, while onboard autonomy relies solely on deterministic BT control.
\end{enumerate}

The remainder of the paper is structured as follows: Section~\ref{sec:related} reviews related work. Section~\ref{sec:architecture} presents the system architecture. Section~\ref{sec:evaluation} describes the experimental setup and results. Section~\ref{sec:conclusion} concludes the paper.

\section{Related Work}
\label{sec:related}

\subsection{Large Language Models for Robot Mission Planning}

LLMs have shown promise for robotic planning through their ability to translate NL instructions into structured task representations. Early frameworks such as Code-as-Policies \citep{liang2022code} and SPINE \citep{ravichandran2025spine} demonstrate that LLMs can synthesize executable or partially specified plans from NL goals. In the maritime domain, AI Captain adapts chain-of-thought reasoning for ship navigation \citep{christensen2025ai}, while \citet{khorrambakht2025composable} combines LLMs with human-in-the-loop decision making to support ship-hull inspection missions.

LLM-assisted BT generation has also been explored. BTGenBot shows that lightweight LLMs can generate syntactically valid Behaviour Trees for single-robot tasks \citep{izzo2024btgenbot}, and LLM-as-BT-Planner improves BT synthesis through prompt engineering \citep{ao2025llm}. However, these approaches rely on LLMs producing low-level BT structures, making validation difficult and remain limited to single-robot settings under reliable communication.

In contrast, our work employs the LLM only for high-level semantic reasoning and multi-vehicle task allocation. By avoiding low-level BT synthesis and validating all LLM outputs through a MILP planner, we simplify interpretation, ensure feasibility, and address the challenges posed by heterogeneous capabilities and inter-task dependencies in multi-AUV missions.

\subsection{Behaviour Trees for Mission Management}

BTs are widely used in robotics due to their modularity, interpretability, and reactive execution model.
Their theoretical foundations and execution semantics were formally established by Colledanchise and Ögren, who also demonstrate their advantages over finite-state machines \citep{colledanchise2018behavior}.

Sprague et al.\ applied BTs in underwater robotics, developing modular and robust control architectures for AUVs and demonstrating their suitability for mission-critical maritime applications characterized by communication dropouts and unpredictable environmental conditions \citep{sprague2018petter}.

More recently, LLMs have been combined with BTs for adaptive mission management. Merino-Fidalgo et al.\ integrate LLM-based reasoning with BT execution for social robots through a Failure Interpreter that diagnoses issues and proposes alternative plans \citep{merino2025behavior}. However, these approaches remain limited to single-robot systems under reliable communication and do not address distributed multi-vehicle settings.

Our work preserves the determinism and verifiability of BT-based execution on each vehicle while enabling adaptivity through topside LLM-assisted reasoning and planning. Rather than modifying the BT structure during operation, we update the task queues consumed by each vehicle BT Mission Manager, ensuring predictable and safe behaviour for critical multi-AUV operations.

\subsection{Multi-Robot Coordination and Shared Autonomy}
Multi-robot coordination in underwater environments is constrained by bandwidth-limited acoustic communication, requiring distributed algorithms that tolerate latency and packet loss.
Communication-aware formation control for AUVs enables teams to adaptively scale coverage while maintaining connectivity \citep{hoff2024communication}.

Failure handling has also been investigated in the context of underwater multi-robot systems.
Failure-aware coordination frameworks model both intermittent and permanent failures and allow teams to switch between centralized and decentralized strategies depending on network availability \citep{li2025failure}.

However, existing multi-vehicle coordination approaches either assume missions specified through structured interfaces (e.g., waypoint files, task matrices) or focus primarily on algorithmic aspects without addressing the operator-level challenge of specifying complex missions in natural language.

\section{System Architecture}
\label{sec:architecture}

\begin{figure}[!t]
\centering
\includegraphics[width=0.95\columnwidth]{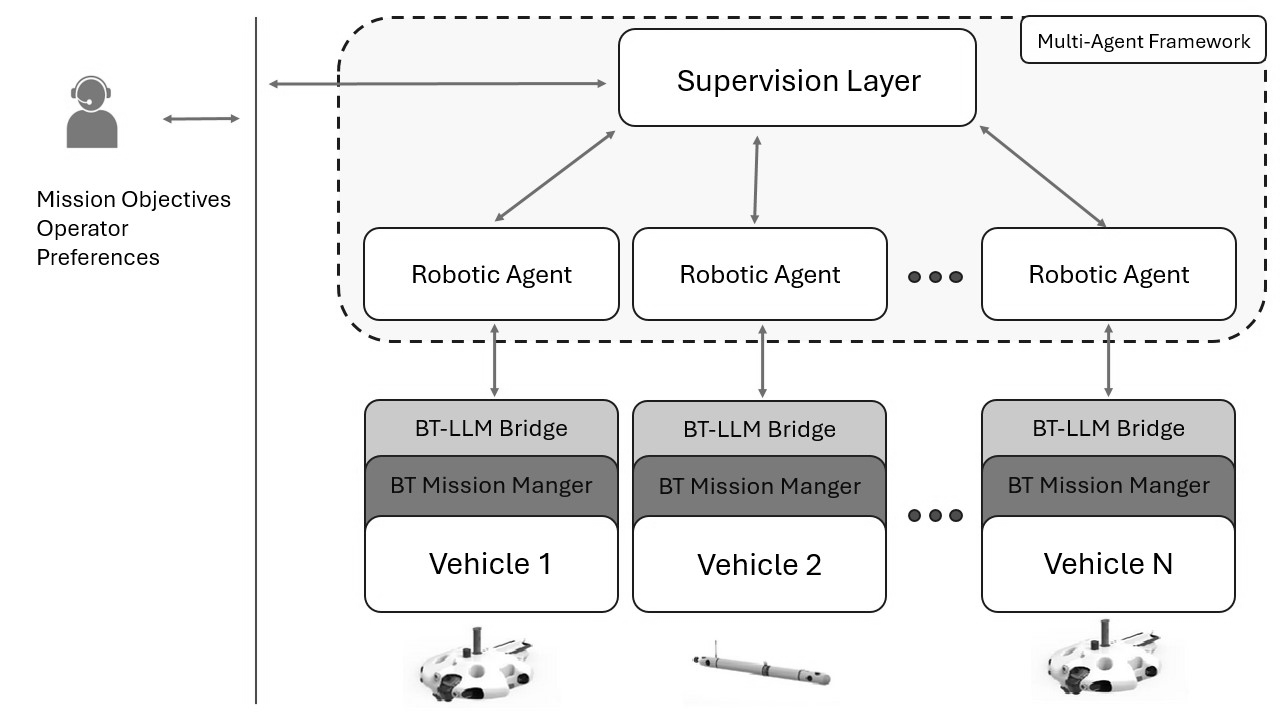}
\caption{Overview of the multi-agent mission-planning architecture, showing the centralized Supervision Layer and per-vehicle Robotic Agents connected to BT Mission Managers via the BT–LLM Bridge.}
\label{fig:architecture}
\end{figure}

The proposed framework integrates LLMs with BTs for multi-robot mission planning and management in maritime environments. As illustrated in Fig.~\ref{fig:architecture}, the architecture follows a centralized reasoning and distributed execution paradigm. It is organized into three main layers:
\begin{itemize}
    \item Multi-Agent Framework for semantic reasoning and mission coordination,
    \item BT-LLM Bridge that translates between topside and vehicles,
    \item Behaviour Tree execution layer for vehicle control.
\end{itemize}

This design separates topside reasoning and validation from onboard execution. LLMs are confined to the planning phase, where they interpret operator intent and extract goals and constraints, while BTs provide robust real-time mission management on each vehicle. Validated mission plans are distributed to individual robots, allowing them to continue executing task sequences even under intermittent or lost communication.

\subsection{Multi-Agent Framework}
\label{subsec:multi}

The proposed agentic framework follows a multi–agent software design in which several coordinated agents interact through structured messages exchanged over a Redis-based communication layer \citep{redis}. The Multi-Agent Framework is implemented using LangGraph \citep{langraph}, and organized into two layers: (i) a \emph{Supervision Layer}, responsible for operator interaction, plan generation, and consistency, and (ii) a \emph{RAs Layer}, where one RA is instantiated per vehicle to manage local execution and feedback. Agents in the Supervision Layer make use of an LLM to interpret operator instructions and extract structured constraints, while safety–critical decisions remain governed by deterministic validation and optimization processes.

\subsubsection{Supervision Layer}

The Supervision Layer comprises two agents: the \emph{Supervisor} and the \emph{Planner}. The Supervisor is the entry point for operator interaction, receiving mission objectives and task constraints in natural language. Agents exchange structured messages over Redis using the format,

\begin{lstlisting}[breaklines=true, columns=flexible, basicstyle=\ttfamily\small]
{"sender": [Agent_ID], "msg_type": [Msg_Type], "payload": {[Msg_Payload]}}
\end{lstlisting}

The Planner interprets operator intent and interfaces with the MILP optimizer. After receiving an NL instruction, it identifies explicit task–robot assignments, ordering relations, or exclusions and extracts them into MILP-compatible constraint structures. Each extracted constraint is validated against the mission specification (task identifiers, robot identifiers, and required sensors). The Planner then calls the MILP optimization tool with the validated constraints and the mission parameters. The resulting capability-aware, time-consistent task allocation is returned to the Supervisor for operator approval.

\paragraph*{MILP Optimizer Tool}
Let \(R\) denote the set of robots, \(T\) the set of mission tasks (waypoints, lawn-mower surveys, spiral surveys), and \(V\) the set of associated nodes. The decision variables include:
\begin{align*}
x_{r,t} &\in \{0,1\} && \text{robot--task assignment}, \\
y_{r,i,j} &\in \{0,1\} && \text{routing between nodes}, \\
A_{r,i} &\ge 0        && \text{arrival times}, \\
D_{r,i} &\ge 0        && \text{departure times}.
\end{align*}
Feasibility is imposed through robot capability constraints, flow-conservation over \(V\), and temporal consistency linking travel and execution times. The objective is to minimize:
\[
\min \; C_{\max} \;+\; \lambda \sum_{r \in R}\sum_{i,j \in V} d_{i,j} \, y_{r,i,j},
\]
where \(C_{\max}\) denotes makespan, \(d_{i,j}\) is the inter-node distances and \(\lambda\) is a tunable weight. The optimization problem is formulated and solved using Gurobi software \citep{gurobi}. NL constraints (e.g., “allocate task~2 to AUVs3” or “perform task~13 before task~2 with AUVs1”) are injected as equality constraints on \(x_{r,t}\) or \(y_{r,i,j}\).

\begin{figure*}[!t]
    \centering
    \includegraphics[width=0.9\linewidth]{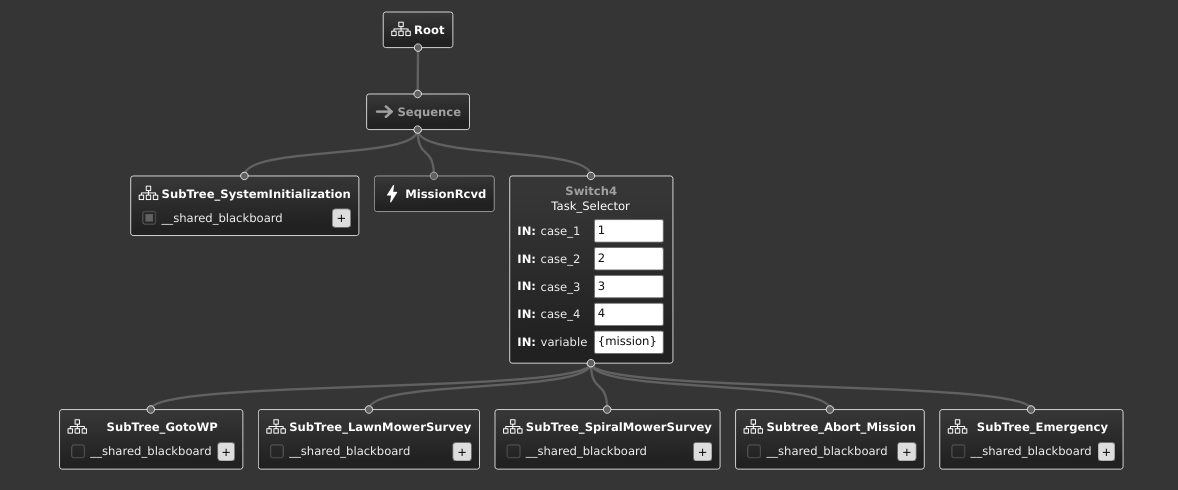}
    \caption{Overview of the BT Mission Manager. The main BT triggers task subtrees. \footnotesize Tree created with Groot~\citep{GrootBT}.}
    \label{fig:mm}
\end{figure*}

\subsubsection{RAs Layer}

Each RA in the \emph{RAs Layer} maintains local execution state (assigned tasks, current task, health). Upon receiving the approved mission plan, the RA retrieves BT execution contexts from the mission database. Each context contains a BT-ready payload specifying task type, waypoint or survey geometry, and required sensors, which the agent forwards to the BT–LLM Bridge for execution.

RAs subscribe to robot-level fault messages over Redis. \emph{Soft failures} correspond to the loss of a sensor. When a soft failure message indicating loss of sensor \(s_{\text{lost}}\) is received, the agent queries the mission task table in the database to determine which remaining tasks require this sensor:
\[
\mathcal{T}_{\text{aff}} = \{ t \in \text{AssignedTasks}(r) : S(t) = s_{\text{lost}} \}.
\]

If the current task does not require \(s_{\text{lost}}\), execution continues. Tasks in \(\mathcal{T}_{\text{aff}}\) are marked as \emph{pending} in the mission database and reported to the Supervisor for reassignment, allowing the robot to operate in degraded mode on feasible tasks.

\emph{Hard failures} are externally triggered events indicating that the vehicle cannot continue the mission. Upon receiving such an event, the \emph{RA} sets its health state to \texttt{CRITICAL}, aborts any ongoing task, and issues a return-to-home command via the BT–LLM Bridge. It then notifies the Supervisor, which initiates centralized replanning to reassign the remaining tasks originally allocated to the failed vehicle.

\emph{RAs} do not coordinate directly with one another. They perform local execution and failure assessment, while all multi-robot decisions—including task allocation, reassignment after failures, and maintenance of global consistency—remain centralized in the Supervision Layer. In a real deployment, RA-to-Supervisor notifications (such as \texttt{REQUEST\_ASSISTANCE} messages) would require the vehicle to surface to establish a communication link with the topside system. In our simulation environment, this communication step is abstracted and implemented as a direct Redis message.

\subsection{BT–LLM Bridge and Execution Layer}

The BT–LLM Bridge is a lightweight component deployed alongside the Execution Layer. Its role is to translate between topside messages and onboard control. The bridge receives task sequences from the corresponding RA, parses the payload, and enqueues the tasks for the BT Mission Manager in the specified order. The bridge forwards task-completion and failure events, together with summarized telemetry and health information to the RA.

The Execution Layer runs on each vehicle and is implemented through a BT Mission Manager \citep{behaviortreecpp}. Each Mission Manager executes a main BT, shown in Fig.~\ref{fig:mm}, structured around a top-level Sequence node.
After successful initialization (\texttt{SubTree\_SystemInitialization}) and mission-reception checks (\texttt{MissionRcvd}), a Task Selector reads the current task identifier from the mission queue and activates the corresponding subtree. Activation follows standard BT tick propagation: at each cycle, the selector evaluates the task identifier and propagates the tick only along the branch matching that identifier. This mechanism ensures that only the subtree relevant to the active task is executed.

Each subtree is composed of:
\begin{itemize}
    \item \emph{precondition checks}, implemented as condition nodes evaluating sensor or state requirements;
    \item \emph{low-level control primitives}, implemented as action nodes interfacing with the vehicle state machines;
    \item \emph{success criteria}, defined by termination conditions such as waypoint arrival;
    \item \emph{recovery logic}, implemented through fallback nodes that attempt retries or alternative actions when failures occur \citep{marzinotto2014towards}.
\end{itemize}

BTs naturally support modularity and reusability, since each subtree encapsulates behaviour that can be recombined or extended without affecting the rest of the tree \citep{iovino2022survey}.
Moreover, BT semantics (\texttt{SUCCESS}, \texttt{FAILURE}, \texttt{RUNNING}) provide explicit task-level status information: on each tick, leaves report their execution state upward, allowing the BT–LLM Bridge to detect task completion, blocking conditions, or execution errors and to trigger replanning when necessary.

\section{Evaluation}\label{sec:evaluation}

This section presents the experimental validation of the proposed Multi-agent software architecture and BT execution framework. Two core hypotheses were evaluated:

\textbf{H1 – Constraint Extraction Accuracy:} the ability of the Planner agent, via its LLM capability, to extract correct task–robot assignments and ordering constraints from NL operator input.

\textbf{H2 – Robustness to Failure:} the ability of the hierarchical framework to maintain mission execution in the presence of robot-level faults through selective replanning and distributed BT execution.

Experiments were conducted across three mission sizes: \emph{small} (2 robots, 8 tasks), \emph{medium} (3 robots, 12 tasks), and \emph{large} (5 robots, 20 tasks). Tasks consisted of waypoint navigation, lawn-mower surveys, and spiral surveys. A task was considered complete once the robot reached the designated waypoint or fully covered the associated survey area. Because the focus of this work is mission-level planning and fault handling, environmental disturbances such as currents, acoustic noise, and localization drift were omitted.

The system was implemented and evaluated in a ROS-based simulator with Zeno AUV kinematics \citep{gelli2018development}. Vehicles traveled at speeds between 1–2~m/s and had heterogeneous sensing capabilities (sonar, optical, sidescan). Robot and sensor faults were injected by sending a \texttt{FAILURE\_EVENT} message over Redis to the corresponding RA. To ensure faults impacted ongoing execution, soft failures (sensor loss) and hard failures (critical robot malfunction) were triggered after a predetermined simulation time, using a fixed simulator time multiplier of 30.

Standard evaluation metrics included Precision, Recall, F1 score, and Constraint Satisfaction Rate (CSR) for H1, and Task Completion Rate (TCR), makespan, total travel distance, and recovery time for H2.

\subsection{H1 – Natural Language Constraint Extraction}

We evaluated constraint extraction using 27 test cases (12 simple, 15 complex) designed to cover single-constraint commands, negations, and multi-constraint compositions for multi-vehicle missions. The Planner agent converts free-form operator instructions into structured MILP constraints using its embedded LLM. For example, the instruction \emph{“Allocate TASK\_2 to AUVs3; Execute TASK\_13 then TASK\_4 with AUVs1”} is processed by the Planner into the assignment and routing constraints:

\begin{lstlisting}[breaklines=true, columns=flexible, basicstyle=\ttfamily\small]
{ "type": "assignment", "robot": "AUVs3", "task": "TASK_2", "value": 1 }
{ "type": "routing", "robot": "AUVs1", "from_node": "TASK_13", "to_node": "TASK_4", "value": 1 }
\end{lstlisting}

which translate directly to constraints on $x_{r,t}$ and $y_{r,i,j}$ in the MILP.

\begin{figure}[!t]
\centering
\includegraphics[width=1\columnwidth]{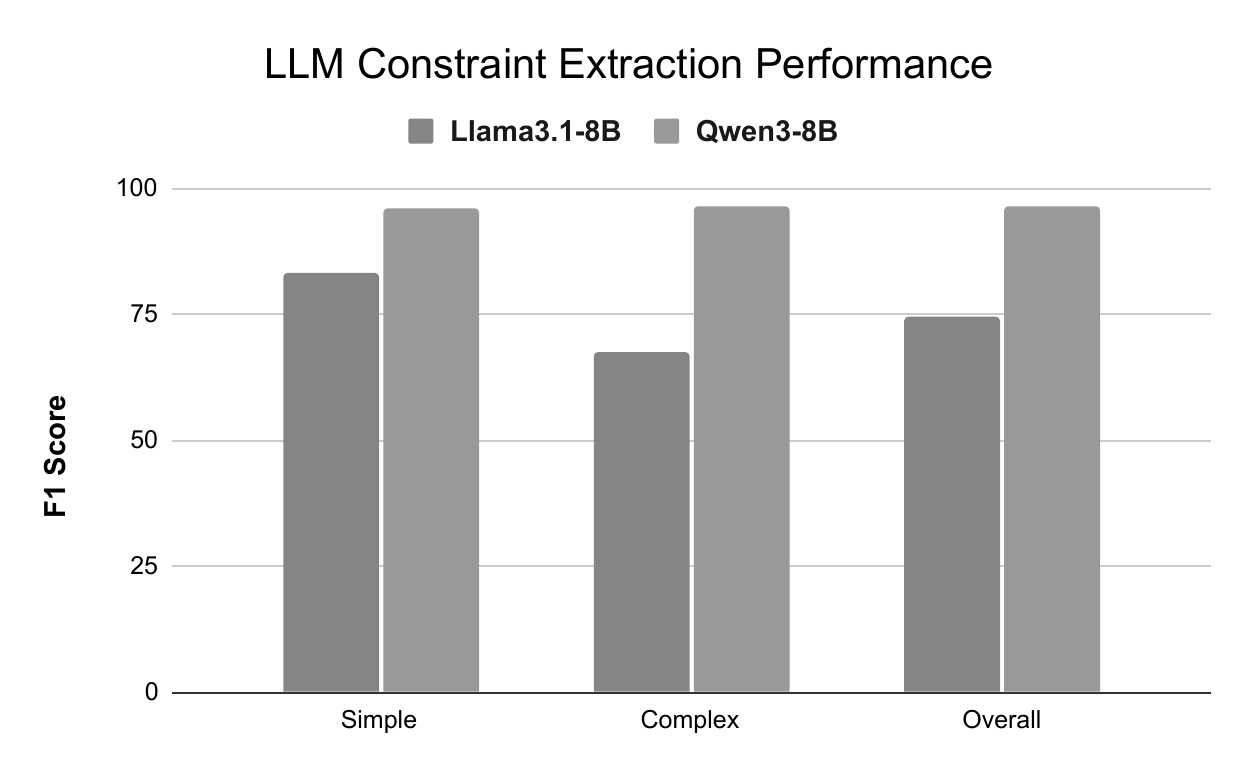}
\caption{F1 scores for the Llama-3.1-8B and Qwen-3-8B models on simple, complex, and overall NL constraint extraction tasks.}
\label{fig:llm_f1}
\end{figure}

Two open-weight LLM models were tested: Llama-3.1-8B and Qwen-3-8B. These models were deployed locally via Ollama \citep{ollama}, and chosen because they can generate structured JSON outputs on commodity hardware, simplifying downstream parsing into MILP-compatible constraints. As shown in Fig.~\ref{fig:llm_f1} and Table~\ref{table:llm_performance}, Qwen-3-8B achieved higher Precision, Recall, and F1 scores across both simple and complex cases. The results indicate that for our specific constraint-extraction patterns used in this application, Qwen produced more reliable structured outputs than Llama.

CSR denotes the percentage of scenarios in which all user-specified constraints are correctly captured in the MILP formulation. For feasible scenarios only, CSR reached 100\% on both models.

\subsection{H2 – Failure-Handling and Hierarchical Recovery}

Fault-handling experiments evaluated four profiles: \emph{Nominal} (no failures), \emph{Soft} (sensor loss), \emph{Hard} (critical robot fault), and \emph{Mixed} (sensor loss followed by a critical fault). Soft failures represent partial capability degradation, whereas hard failures permanently remove a robot from the mission. Failures were injected mid-mission by sending a \texttt{FAILURE\_EVENT} message over Redis to the corresponding RA, ensuring that the affected robot was actively executing or scheduled to execute tasks when the fault occurred. Upon receiving a failure event, the RA applies the response rules described in Sec.~\ref{subsec:multi} and issues a \texttt{REQUEST\_ASSISTANCE} message to the Supervisor agent when replanning is required.

Across all failure profiles and mission sizes, the framework achieved a 100\% TCR, indicating that hierarchical replanning combined with distributed BT execution is sufficient to preserve mission completion despite robot-level faults. For small missions, the mean recovery time (elapsed time between a \texttt{REQUEST\_ASSISTANCE} message and distribution of a new task allocation) was 8.2~s, with scenario averages of 11.3~s (Soft), 1.8~s (Hard), and 11.5~s (Mixed). For medium missions, the mean recovery time was 13.4~s, with scenario averages of 12.5~s (Soft), 12.4~s (Hard), and 15.1~s (Mixed). These values reflect the time required to recompute and redistribute plans rather than the type of failure, since replanning cost scales more with mission size than with failure mode.

\begin{figure}[!h]
\centering
\includegraphics[width=1\columnwidth]{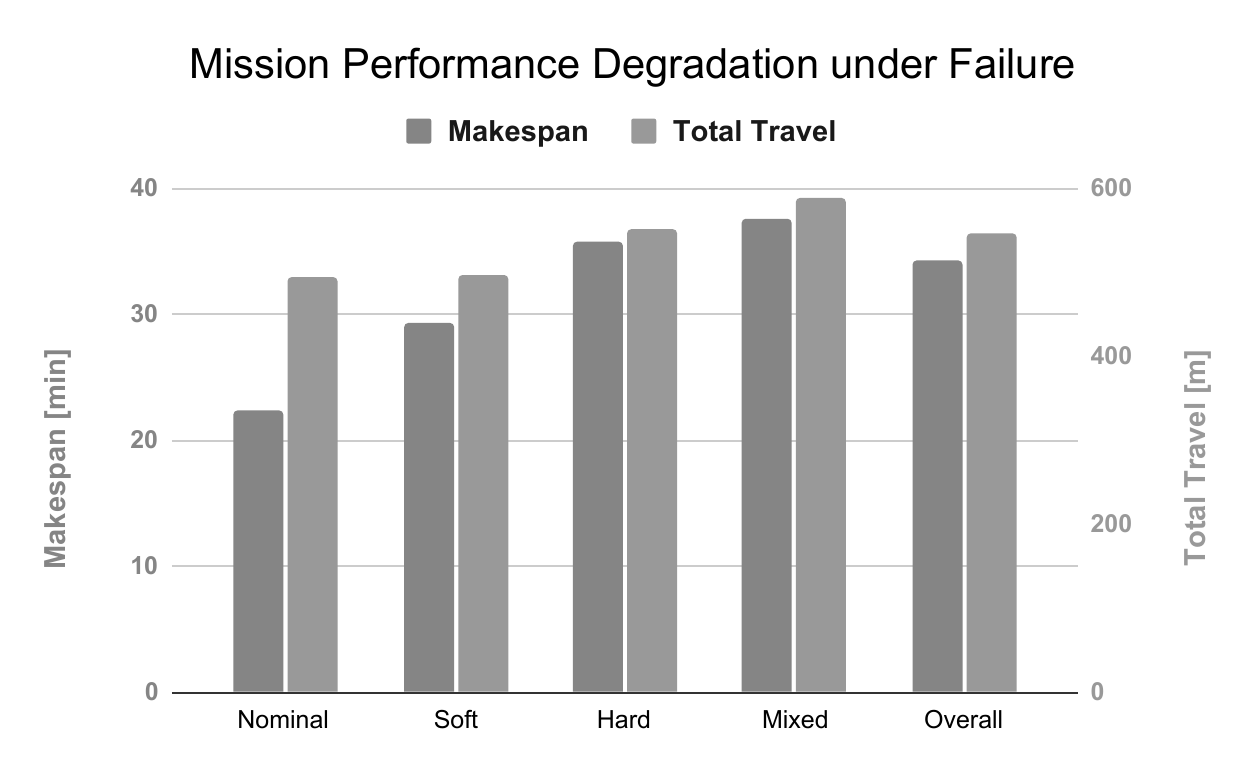}
\caption{Mission-level effects of soft, hard, and mixed failures on makespan and total travel distance for the small- and medium-scale missions (mean values).}
\label{fig:missiondeg}
\end{figure}

Fig.~\ref{fig:missiondeg} illustrates the increases in makespan and total travel distance under soft, hard, and mixed failures for both mission scales. These increases are consistent with the additional path adjustments and reassignment decisions introduced during recovery.

Communication overhead remained moderate because RAs communicate only fault-related events and do not exchange information with one another. In all evaluated experiments, a fault triggered only a small number of coordination messages (typically three): one failure notification from the robot, one replanning request, and one updated task distribution from the Supervisor. While we do not compare against a specific centralized baseline, these absolute message counts characterize the communication footprint of the proposed architecture and reflect its ability to localize failure handling while maintaining consistency.

\begin{table}[!t]
\caption{Performance of LLM models in NL constraint extraction.}
\centering

\resizebox{\columnwidth}{!}{
\begin{tabular}{|c|c|c|c|c|c|}
\hline
\textbf{Model} & \textbf{Complexity} & \textbf{Precision} & \textbf{Recall} & \textbf{F1} & \textbf{CSR} \\ \hline

\multirow{3}{*}{\textbf{Llama3.1:8B}}
 & Simple  & 83.3 & 83.3 & 83.3 & 41.7 \\ \cline{2-6}
 & Complex & 67.2 & 70.0 & 67.4 & 81.7 \\ \cline{2-6}
 & Overall & 74.4 & 75.9 & 74.5 & 63.9 \\ \hline

\multirow{3}{*}{\textbf{Qwen3:8B}}
 & Simple  & 94.4 & 100.0 & 95.8 & 49.4 \\ \cline{2-6}
 & Complex & 96.6 & 96.6 & 96.6 & 92.8 \\ \cline{2-6}
 & Overall & 95.6 & 98.1 & 96.2 & 73.5 \\ \hline

\end{tabular}
}
\par\smallskip\parbox{\columnwidth}{\footnotesize \begin{math}\text{CSR} < 100\%\end{math} occurs when user specified infeasible constraints.
}

\label{table:llm_performance}
\end{table}

\section{Conclusion}
\label{sec:conclusion}

In this paper, we introduced a hierarchical mission-planning architecture that combines LLM-based interpretation of NL instructions with deterministic BT execution for multi-AUV operations. The proposed framework enables operators to specify mission intent at a high semantic level, while the MILP optimizer provides feasible, capability-aware task allocations. Distributed BT execution allows vehicles to progress autonomously during communication gaps, and Robotic Agents support local handling of soft and hard failures by identifying affected tasks and requesting selective replanning.

Experimental results demonstrate high constraint-extraction accuracy, full mission completion under a range of failure scenarios, and moderate communication requirements due to the localized nature of failure reporting and selective task redistribution. These findings indicate that the proposed integration of LLM-based intent interpretation, formal validation, and BT-controlled execution can enhance multi-vehicle autonomy in bandwidth-limited underwater environments.

Future work will focus on validating the framework with real AUV platforms, expanding the natural-language interface to support richer task semantics, and investigating the integration of lightweight LLM capabilities within Robotic Agents as onboard computational resources permit.

\begin{ack}
This work has been supported in part by the Italian Ministry of Education and Research (MIUR) in the framework of the FoReLab project (Departments of Excellence) and by MUR through the Project PRIN COMET CUP: 2022TPSX25 (EU NextGenerationEU plan via the Italian Bando Prin 2022 - D.D. 104 del 02-02-2022).
\end{ack}

\section*{DECLARATION OF GENERATIVE AI AND AI-ASSISTED TECHNOLOGIES IN THE WRITING PROCESS}
During the preparation of this work, the author(s) used $ChatGPT (OpenAI)$ for assistance with language refinement, organization of paragraphs, and improvement of clarity and flow. After using this tool/service, the author(s) reviewed and edited the content as needed and take(s) full responsibility for the content of the publication.

\bibliography{ifacconf}

\end{document}